\documentclass[letterpaper]{article} 
\usepackage{aaai2027}  
\usepackage[hyphens]{url} 
\usepackage{graphicx} 
\usepackage{natbib}  
\usepackage{caption} 
\usepackage{algorithm}
\usepackage{algorithmic}
\usepackage{multirow}

\usepackage{newfloat}
\usepackage{listings}
\DeclareCaptionStyle{ruled}{labelfont=normalfont,labelsep=colon,strut=off} 
\floatstyle{ruled}
\newfloat{listing}{tb}{lst}{}
\floatname{listing}{Listing}

\usepackage{booktabs}
\usepackage{amssymb}

\newcommand{\cmark}{\ensuremath{\checkmark}}

\usepackage{amsmath}
\usepackage{amssymb}
\usepackage{microtype}
\title{AdaAnchor4D: Anchor-Conditioned Spatiotemporal Feature Aggregation for Monocular UAV 4D Reconstruction}

\author {
    Peiyi Xu\textsuperscript{\rm 1},
    Junpeng Zhang\textsuperscript{\rm 1},
    Guanbin Li\textsuperscript{\rm 2},
    Ronghua Shang\textsuperscript{\rm 1},
    Mingtao Feng\textsuperscript{\rm 1},
    Le Dong\textsuperscript{\rm 1},
    Weisheng Dong\textsuperscript{\rm 1},
    Guangming Shi\textsuperscript{\rm 1},
    Jie Feng\textsuperscript{\rm 1}\corresponding
}
\affiliations{
    \textsuperscript{\rm 1}Xidian University, Xi'an, China\\
    \textsuperscript{\rm 2}Sun Yat-sen University, Guangzhou, China\\
}

\begin{document}

\maketitle

\begin{abstract}
Monocular UAV videos provide valuable observations for dynamic reconstruction of complex urban scenes. However, such scenes exhibit pronounced spatiotemporal heterogeneity: different regions follow distinct temporal activity patterns, while the motion states of some dynamic regions may further evolve over time. Although dynamic Gaussian methods based on decomposed shared spatiotemporal feature fields have achieved efficient and accurate reconstruction in object-centric or relatively compact scenes, their commonly adopted fixed plane-wise feature combination mechanisms are less suited to the heterogeneous local dynamics of UAV scenes, often leading to ghosting artifacts and blurred dynamic details.
To address this challenge, we propose AdaAnchor4D, an adaptive anchor
deformation framework for monocular UAV dynamic scene reconstruction. At its
core, Anchor-Conditioned Feature Aggregation (ACFA) adaptively aggregates
shared spatiotemporal features using anchor-specific aggregation embeddings
and temporal information, allowing different local units to obtain dynamic
representations tailored to their local and temporal states. Decoupled Local
Geometry Deformation (DLGD) separates anchor-state deformation from local
Gaussian geometry deformation, while Density-Adaptive Coordinate Warping
(DACW) reparameterizes feature-query coordinates according to the axis-wise
anchor distributions, alleviating the mismatch between non-uniform geometric
sampling and uniform grid parameterization.
Experiments on UAV-Arc4D, VisDrone, and UAVDT show that AdaAnchor4D achieves
higher rendering quality than representative dynamic Gaussian methods while
maintaining real-time rendering performance. The code will be made publicly
available.

\end{abstract}

\begin{figure}[t]
    \centering
    \includegraphics[width=0.95\linewidth]{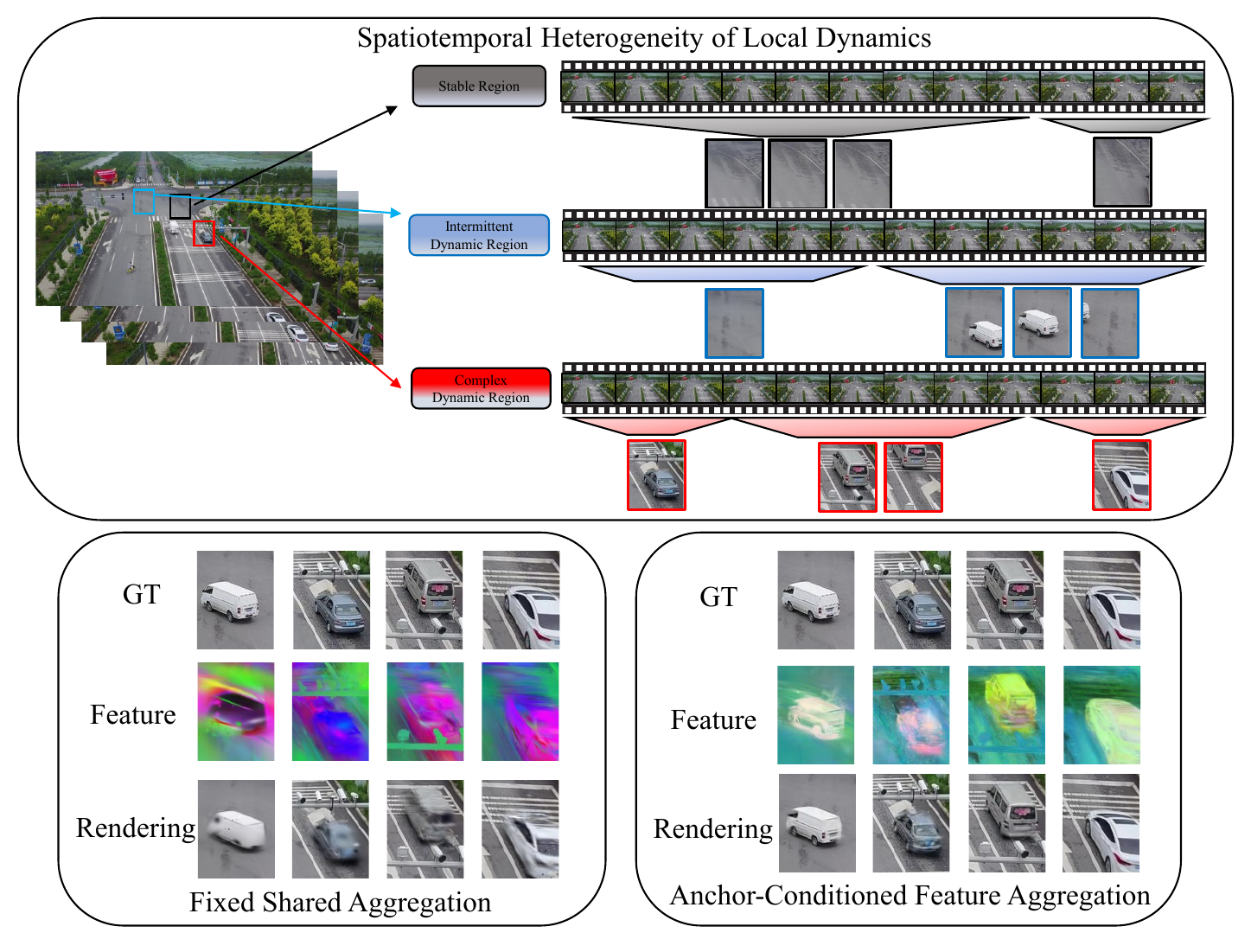}
    \caption{Monocular UAV scenes exhibit strong spatiotemporal heterogeneity across regions and time. Fixed aggregation of shared spatiotemporal features cannot adapt to varying local dynamics, often causing blur and ghosting. Our Anchor-Conditioned Feature Aggregation uses anchor-specific embeddings and temporal cues to adaptively reweight shared features for improved dynamic modeling.}
    \label{fig:1}
\end{figure}

\section{Introduction}
Recent advances in 3D Gaussian Splatting (3DGS)~\cite{kerbl3Dgaussians} have demonstrated remarkable performance in high-quality novel-view synthesis, benefiting from its explicit scene representation and efficient differentiable rendering. Building upon 3DGS, dynamic Gaussian splatting methods further introduce the temporal dimension, providing a new paradigm for continuous spatiotemporal representation and real-time rendering of dynamic scenes. Monocular UAV videos offer flexible viewpoints, wide-area coverage, and convenient data acquisition, providing valuable observations for dynamic reconstruction of complex urban scenes. Extending dynamic Gaussian splatting to monocular UAV videos is therefore a natural step toward its deployment in large-scale and complex urban environments.

However, such scenes exhibit pronounced spatiotemporal heterogeneity. As illustrated in Fig.~\ref{fig:1}, different regions follow distinct temporal activity patterns throughout a video sequence: some remain stable over extended periods, some change only within limited time intervals, while others undergo persistent and complex motion. Meanwhile, the motion state of the same dynamic region may also evolve over time. Consequently, different local units require distinct dynamic representations at different time steps. How to construct dynamic representations adapted to local scene conditions and temporal states while maintaining overall modeling efficiency thus becomes an important challenge for dynamic UAV scene reconstruction.

To balance representational capacity and computational efficiency, many dynamic Gaussian methods adopt canonical-space deformation frameworks and model scene dynamics with decomposed shared spatiotemporal fields~\cite{fridovich2023k}. These methods compactly encode high-dimensional spatiotemporal information using multiple low-dimensional feature planes and aggregate the queried plane features according to predefined rules. Such shared modeling is compact and efficient, and has achieved strong performance in object-centric or relatively compact scenes. However, fixed feature combination rules cannot adapt the contributions of different planes and channels to individual local units and their temporal states, limiting the flexibility of shared spatiotemporal representations in modeling heterogeneous local dynamics in UAV scenes and potentially leading to ghosting artifacts and blurred dynamic details in regions with complex motion. Therefore, shared spatiotemporal fields should preserve their compact representation while enabling adaptive feature aggregation conditioned on different local units and their temporal states.

To this end, we propose AdaAnchor4D, a monocular UAV 4D reconstruction framework centered on anchor-conditioned spatiotemporal feature aggregation. AdaAnchor4D employs anchors as compact units for local dynamic modeling and constructs locally adaptive dynamic representations conditioned on anchor-specific aggregation embeddings and temporal information, while retaining the efficiency of shared spatiotemporal representations. Specifically, Anchor-Conditioned Feature Aggregation predicts channel-wise plane aggregation weights based on anchor-specific aggregation embeddings and temporal information, adaptively modulating the contributions of different shared spatiotemporal features. This enables different local units at different time steps to extract representations from the same shared spatiotemporal field that are tailored to their local dynamic requirements. ACFA preserves a unified shared spatiotemporal representation and only conditions the feature aggregation process, thereby enhancing its adaptability to heterogeneous local dynamics while maintaining representation compactness.

Building upon ACFA, we further observe that jointly modeling deformation variables at different representation levels may limit the flexibility of fine-grained geometry modeling. We therefore introduce Decoupled Local Geometry Deformation, which separates ACFA-driven anchor-state deformation from local geometry residual prediction and provides dedicated modeling pathways for deformations at different levels. In addition, due to UAV imaging geometry and scene structure, anchors are often distributed non-uniformly in space, which is mismatched with the uniform coordinate parameterization of shared feature grids. We thus propose Density-Adaptive Coordinate Warping, which reparameterizes query coordinates according to the axis-wise anchor distributions, allowing the limited grid capacity to better accommodate non-uniform geometric sampling.

Experiments on UAV-Arc4D, VisDrone, and UAVDT demonstrate that AdaAnchor4D
improves rendering quality while maintaining real-time rendering performance
and a compact neural representation.

Our main contributions are summarized as follows:

\begin{itemize}
\item We propose AdaAnchor4D, an adaptive anchor deformation framework for monocular UAV 4D reconstruction. At its core, Anchor-Conditioned Feature Aggregation adaptively aggregates shared spatiotemporal features using anchor-specific aggregation embeddings and temporal information.

\item We introduce Decoupled Local Geometry Deformation (DLGD) to separate anchor-state deformation from local Gaussian geometry deformation, enabling more flexible local geometry modeling.
\item We propose Density-Adaptive Coordinate Warping to adapt feature-query coordinates to non-uniform anchor distributions and improve the allocation of shared feature-grid capacity.

\end{itemize}

\section{Related Work}
\subsection{Dynamic Gaussian Splatting}
Dynamic Gaussian Splatting extends 3DGS~\cite{kerbl3Dgaussians} to dynamic scenes through either explicit spatiotemporal modeling or canonical-space deformation. Explicit methods represent temporal evolution using time-varying Gaussian states~\cite{luiten2023dynamic,lee2024fully}, temporal basis functions or parameterized trajectories~\cite{li2024spacetime,lin2024gaussian,kratimenos2024dynmf}, or Gaussian primitives defined directly in four-dimensional spacetime~\cite{duan20244d,yang2024real,zhou20264c4d}. While these representations directly describe Gaussian motion over time, they generally introduce additional temporal parameters or higher-dimensional primitives.

Canonical-space deformation methods instead maintain a base Gaussian representation and predict its time-dependent state through deformation fields~\cite{yang2024deformable,TuYing2025SpeeDe3DGS}. To improve efficiency, many approaches adopt decomposed spatiotemporal fields that store dynamic information in low-dimensional feature planes and aggregate queried features for deformation prediction~\cite{fridovich2023k,wu20244d,kwak2025modec,wang2025degauss,yao2025sd,kim20244d}. However, their plane features are commonly combined using predefined rules that cannot adapt to different local units and temporal states. Grid4D instead decomposes the 4D input into spatial and spatiotemporal 3D hash encodings and uses spatially derived directional attention to modulate temporal features~\cite{xu2024grid4d}. In contrast, ACFA retains a compact multi-plane shared field and predicts channel-wise plane aggregation weights from anchor-specific aggregation embeddings and temporal information.

Several recent methods~\cite{kwak2025modec,yao2025sd,chen2026haif,cho20264d,kwak2026morel} further organize dynamic Gaussians using anchor-based representations inspired by Scaffold-GS~\cite{lu2024scaffold}. Representative recent methods include MoDec-GS, 4D Scaffold Gaussian Splatting (4D-SFGS), and MoRel. MoDec-GS models global anchor deformation followed by local Gaussian deformation~\cite{kwak2025modec}, 4D-SFGS generates local neural 4D Gaussians from grid-aligned 4D anchors~\cite{cho20264d}, and MoRel connects local canonical anchor spaces through bidirectional deformation and temporal blending~\cite{kwak2026morel}. Unlike these methods, AdaAnchor4D uses anchor-specific
aggregation embeddings to adaptively aggregate a shared
spatiotemporal field for anchor-state deformation, while
geometry condition embeddings provide dedicated pathways
for local Gaussian geometry deformation.
\subsection{UAV Scene Reconstruction}
With flexible viewpoints and wide-area coverage, UAVs have been widely adopted for 3D reconstruction of urban environments, roads, and natural scenes. Existing studies mainly focus on static aerial scenes and improve reconstruction quality and efficiency through camera pose refinement, geometric consistency constraints, large-scale scene partitioning and representation, and model compression~\cite{jia2024drone,wu20243d,lin2024vastgaussian,li2026urbangs}.

In contrast, research on dynamic UAV scene reconstruction remains relatively limited. TK-Planes extends the K-Planes representation and models small-scale dynamic objects under high-altitude viewpoints using hierarchical feature vectors~\cite{maxey2025tk}. UAV4D combines 3D foundation models with human priors to jointly recover static backgrounds and dynamic humans from monocular UAV videos containing multiple moving pedestrians~\cite{choi2026uav4d}. AeroGS targets monocular UAV videos with unknown camera poses and jointly recovers camera trajectories and dynamic scene representations, while introducing scale-aware spatiotemporal anchors to alleviate the coupling among camera motion, object motion, and scale variation~\cite{li2026aerogs}. AeroDGS addresses depth and motion ambiguities in monocular aerial settings by introducing monocular geometric lifting and physical consistency constraints, leading to more stable estimation of dynamic geometry and motion~\cite{liu2026aerodgs}.

These studies mainly address dynamic UAV scene reconstruction from the perspectives of human priors, camera trajectory recovery, and geometric or motion constraints for dynamic objects. In contrast, our work focuses on the dynamic scene representation itself, with the goal of improving the adaptability of shared spatiotemporal fields to heterogeneous local dynamics.

\section{Preliminaries}

\paragraph{Anchor-based Gaussian Representation.}
Instead of maintaining each Gaussian primitive independently, an anchor-based
representation~\cite{lu2024scaffold} associates each anchor with $K$ local
Gaussians. Each canonical anchor consists of a center $\mathbf{x}_i$ and a
learnable feature $\mathbf{f}_i$, which represent its anchor-level state,
together with local offsets
$\mathbf{O}_i=\{\mathbf{o}_{i,k}\}_{k=1}^{K}$ and scale parameters
$\mathbf{s}_i\in\mathbb{R}^{6}$, which characterize the Gaussian-level
geometry within the anchor. The first three components of $\mathbf{s}_i$
scale the local offsets, and each Gaussian center is obtained by adding the
scaled offset to the anchor center, while the remaining three components
provide covariance-related scaling. The opacity, view-dependent color,
rotation, and final anisotropic scale of each local Gaussian are decoded from
the learnable anchor feature $\mathbf{f}_i$ together with the camera-to-anchor
viewing information, following the standard anchor-based representation.

\paragraph{Spatiotemporal Feature Factorization.}
As in 4D-GS~\cite{wu20244d},
the shared spatiotemporal feature field is decomposed into six learnable 2D
feature planes, instead of being represented as a dense feature grid in
$(x,y,z,t)$. These planes correspond to all pairwise coordinate combinations:
$
\mathcal{P}=\{xy,xz,yz,xt,yt,zt\}.
$
At resolution level $l$, the feature grid on plane $p\in\mathcal{P}$ is denoted by
$
\mathbf{H}_{p}^{(l)}
\in
\mathbb{R}^{C\times R_{p,1}^{(l)}\times R_{p,2}^{(l)}},
$
where $C$ is the feature dimension, and $R_{p,1}^{(l)}$ and
$R_{p,2}^{(l)}$ are the corresponding plane resolutions.

Given a spatiotemporal query coordinate $\mathbf{q}=(x,y,z,t)$, its feature
on plane $p$ is obtained through bilinear interpolation:
\begin{equation}
\mathbf{g}_{p}^{(l)}(\mathbf{q})
=
\operatorname{Interp}
\left(
\mathbf{H}_{p}^{(l)},
\pi_p(\mathbf{q})
\right),
\end{equation}
where $\pi_p(\cdot)$ projects the 4D coordinate onto plane $p$.
The resulting plane-wise features are aggregated through element-wise
multiplication:
\begin{equation}
\mathbf{h}^{(l)}(\mathbf{q})
=
\prod_{p\in\mathcal{P}}
\mathbf{g}_{p}^{(l)}(\mathbf{q}).
\end{equation}
The features from all $L$ resolution levels are then concatenated:
\begin{equation}
\mathbf{h}(\mathbf{q})
=
\operatorname{Concat}_{l=1}^{L}
\left(
\mathbf{h}^{(l)}(\mathbf{q})
\right).
\end{equation}
This fixed aggregation applies the same plane-wise combination rule to all
queries, without adapting the contribution of each plane to different spatial
locations and temporal states.
\section{Method}

\begin{figure*}[t]
    \centering
    \includegraphics[width=1.0\textwidth]{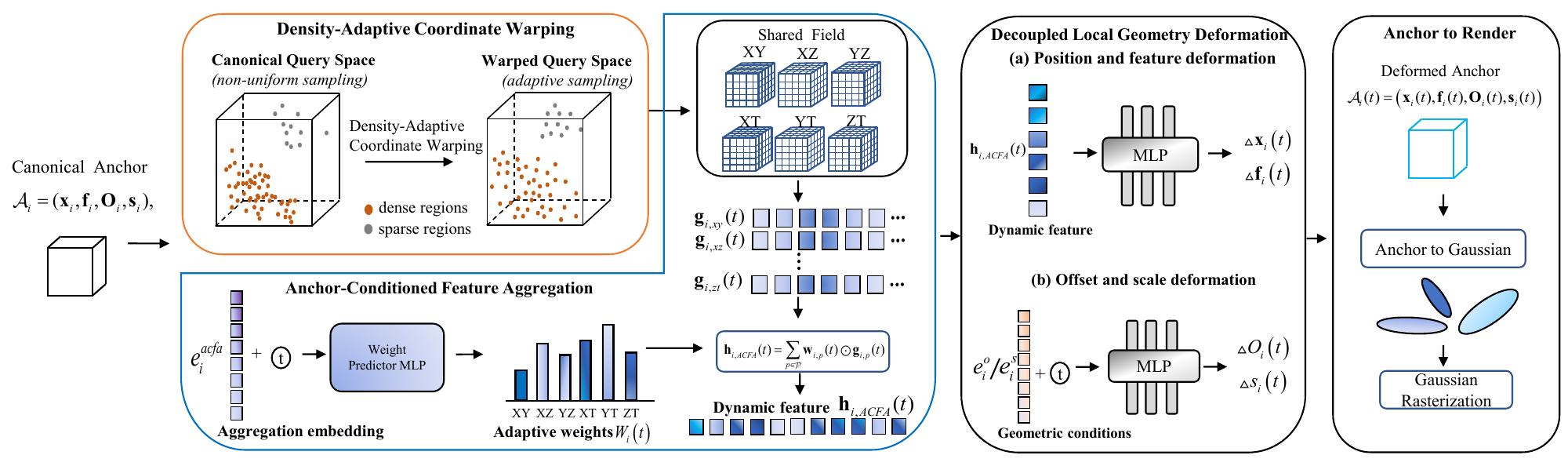}
    \caption{
    Overview of AdaAnchor4D. Anchor-Conditioned Feature Aggregation performs anchor- and time-dependent feature aggregation, Decoupled Local Geometry Deformation handles anchor-state and local Gaussian geometry deformation with separate branches, and Density-Adaptive Coordinate Warping adjusts the query coordinates to the coarse-stage anchor distribution.
}
    \label{fig:2}
\end{figure*}

\subsection{Overview of AdaAnchor4D}
To model heterogeneous dynamics in monocular UAV scenes within an anchor-based
representation, we propose AdaAnchor4D, an
adaptive anchor deformation framework for dynamic scene reconstruction from
monocular UAV videos. As illustrated in Fig.~\ref{fig:2}, AdaAnchor4D consists
of Density-Adaptive Coordinate Warping (DACW),
Anchor-Conditioned Feature Aggregation (ACFA), and
Decoupled Local Geometry Deformation (DLGD).

Given a canonical-space anchor $\mathcal{A}_i$ and a target time $t$, ACFA
predicts channel-wise aggregation weights over the shared feature planes
conditioned on an anchor-specific aggregation embedding and time. The resulting
dynamic representation is used to predict residuals for the anchor position
and feature. In parallel, DLGD uses two geometry condition embeddings and
dedicated branches to predict residuals for the local Gaussian offsets and
scale parameters, while DACW reparameterizes the spatial feature-query
coordinates through nonlinear mappings constructed from the axis-wise anchor
distributions obtained after the coarse stage. The shared plane features used
by ACFA are queried at the resulting warped coordinates.

These residuals deform the canonical anchor into
$\mathcal{A}_i(t)=
\bigl(
\mathbf{x}_i(t),
\mathbf{f}_i(t),
\mathbf{O}_i(t),
\mathbf{s}_i(t)
\bigr)$.
The associated local Gaussians are then instantiated from the deformed anchor
for differentiable rendering.
\subsection{Anchor-Conditioned Feature Aggregation}
Factorized spatiotemporal fields provide a compact representation of scene dynamics through globally shared feature planes. However, fixed aggregation applies the same feature readout rule to all anchors, limiting its ability to model the heterogeneous and evolving local motions commonly observed in urban scenes. To address this limitation, we propose Anchor-Conditioned Feature Aggregation (ACFA), which retains a globally shared spatiotemporal field while allowing each anchor to adaptively aggregate plane features over time.

Specifically, ACFA associates each anchor with a learnable aggregation
embedding $\mathbf{e}_i^{\mathrm{acfa}}\in\mathbb{R}^{C}$, which provides an
anchor-specific condition for plane-feature aggregation.
This embedding is jointly optimized with the other anchor parameters and propagated during anchor densification.  Together with the normalized time $t\in[0,1]$, this embedding is fed into a lightweight predictor $\Phi_w$ to produce channel-wise plane scores:
\begin{equation}
\mathbf{Z}_i(t)=
\Phi_w\left(
[\mathbf{e}_i^{\mathrm{acfa}},t]
\right)
\in
\mathbb{R}^{|\mathcal{P}|\times C}.
\end{equation}

For each channel, the scores are normalized across the plane dimension using softmax\begin{equation}
w_{i,p,c}(t)
=
\frac{e^{Z_{i,p,c}(t)}}
{\sum_{p'\in\mathcal{P}} e^{Z_{i,p',c}(t)}}.
\end{equation}

This yields anchor- and time-dependent plane weights for each feature channel.

The aggregated feature at resolution level $l$ is
\begin{equation}
\mathbf{h}_{i,\mathrm{ACFA}}^{(l)}(t)
=
\sum_{p\in\mathcal{P}}
\mathbf{w}_{i,p}(t)\odot
\mathbf{g}_{i,p}^{(l)}(t),
\end{equation}
where $\mathbf{g}_{i,p}^{(l)}(t)$ denotes the plane feature queried at the
DACW-warped coordinate, $\mathbf{w}_{i,p}(t)$ contains the channel-wise
weights for plane $p$, and $\odot$ denotes element-wise multiplication.
The same weights are shared across resolution levels, and the final feature is obtained as
$\mathbf{h}_{i,\mathrm{ACFA}}(t)
=
\operatorname{Concat}_{l=1}^{L}
\bigl(\mathbf{h}_{i,\mathrm{ACFA}}^{(l)}(t)\bigr)$.

To improve optimization stability, we zero-initialize the output layer of
$\Phi_w$, assigning every plane a uniform weight of $1/|\mathcal{P}|$ at the
beginning of training. ACFA therefore starts from uniform plane weighting and
progressively learns aggregation patterns that vary across anchors and time.
In this way, ACFA preserves the shared spatiotemporal field while allowing
different anchors to adopt differentiated feature-reading patterns for heterogeneous local dynamics.

\subsection{Decoupled Local Geometry Deformation}

ACFA provides adaptive dynamic features for each anchor at each time step. However, the anchor position and feature characterize the anchor-level state,
whereas the local Gaussian offsets and scale parameters describe the intra-anchor geometry of the associated Gaussians. Predicting all these attributes from the same dynamic representation couples deformation variables across different representation levels, potentially limiting the flexibility of local geometry modeling. We therefore propose Decoupled Local Geometry Deformation (DLGD), which retains ACFA-based prediction for anchor-level attributes while introducing dedicated conditioning pathways for local geometric deformation.

Given $\mathbf{h}_{i,\mathrm{ACFA}}(t)$, the residuals of the anchor position
and feature are predicted by a shared dynamic encoder followed by two
attribute-specific heads:
\begin{equation}
\begin{aligned}
\Delta\mathbf{x}_i(t)
&=
\Phi_x\left(
\Phi_{\mathrm{enc}}
\left(\mathbf{h}_{i,\mathrm{ACFA}}(t)\right)
\right),\\
\Delta\mathbf{f}_i(t)
&=
\Phi_f\left(
\Phi_{\mathrm{enc}}
\left(\mathbf{h}_{i,\mathrm{ACFA}}(t)\right)
\right).
\end{aligned}
\end{equation}
The corresponding anchor states are updated as
\begin{equation}
\mathbf{x}_i(t)=\mathbf{x}_i+\Delta\mathbf{x}_i(t),
\qquad
\mathbf{f}_i(t)=\mathbf{f}_i+\Delta\mathbf{f}_i(t).
\end{equation}

For local geometry, we associate each anchor with two learnable geometry
condition embeddings, $\mathbf{e}_i^{o}\in\mathbb{R}^{C}$ and $\mathbf{e}_i^{s}\in\mathbb{R}^{C}$, which
provide dedicated conditions for offset and scale-parameter deformation,
respectively. 
Two lightweight branches predict the offset and scale residuals:
\begin{equation}
\Delta\mathbf{O}_i(t)
=
\Phi_o\left([\mathbf{e}_i^{o},t]\right),
\qquad
\Delta\mathbf{s}_i(t)
=
\Phi_s\left([\mathbf{e}_i^{s},t]\right).
\end{equation}
The local geometry is then updated by
\begin{equation}
\mathbf{O}_i(t)
=
\mathbf{O}_i+\Delta\mathbf{O}_i(t),
\qquad
\mathbf{s}_i(t)
=
\mathbf{s}_i\odot\exp\left(\Delta\mathbf{s}_i(t)\right).
\end{equation}

The output layers of $\Phi_o$ and $\Phi_s$ are zero-initialized, ensuring that optimization starts from the canonical geometry. By decoupling local geometry prediction from ACFA-driven anchor-state deformation, DLGD provides dedicated conditioning pathways for deformation variables at different representation levels, enabling more flexible fine-grained geometry modeling.

\subsection{Density-Adaptive Coordinate Warping}
Anchors in UAV scenes are often distributed non-uniformly along different spatial axes, whereas shared feature grids typically adopt uniform coordinate parameterization, leading to a mismatch between limited feature capacity and the underlying geometric sampling distribution. 
To alleviate this mismatch, we propose Density-Adaptive Coordinate Warping (DACW), which constructs density-adaptive nonlinear query mappings according to the marginal distribution of anchors along each spatial axis.

Given the linearly normalized canonical anchor coordinate
$
\mathbf{u}_i =
(u_{i,x},u_{i,y},u_{i,z}),
$
DACW independently constructs a nonlinear reparameterization mapping for each of the three spatial axes. For an axis $d\in\{x,y,z\}$, we uniformly divide $[-1,1]$ into $B$ intervals. We then construct an axis-wise anchor histogram, apply Gaussian smoothing, and normalize it to obtain the probability mass $p_{d,b}$ of the $b$-th interval, where $\sum_{b=1}^{B} p_{d,b}=1$.

To adapt the warping strength to the local sampling density, we first compute the density response of each interval relative to a uniform distribution:
\begin{equation}
q_{d,b}
=
\left(
\frac{p_{d,b}}{1/B}
\right)^{\gamma},
\end{equation}
where $\gamma$ controls the sensitivity to density variations. The response is then scaled to lie within a predefined range:
\begin{equation}
\lambda_{d,b}
=
\lambda_{\min}
+
(\lambda_{\max}-\lambda_{\min})
\frac{q_{d,b}}
{\max_{b'} q_{d,b'}}.
\end{equation}
where $0 \leq \lambda_{\min} \leq \lambda_{\max} \leq 1$
specify the lower and upper bounds of the density-adaptation strength.

To balance density-adaptive and uniform coordinate allocation, we define an
unnormalized target allocation as
\begin{equation}
r_{d,b}
=
(1-\lambda_{d,b})\frac{1}{B}
+
\lambda_{d,b}p_{d,b},
\quad
\bar r_{d,b}
=
\frac{r_{d,b}}
{\sum_{j=1}^{B} r_{d,j}}.
\end{equation}

Let the input interval boundaries be
$
e_b=-1+\frac{2b}{B},
\forall
b \in \{0,\cdots,B\}
$.
The corresponding output boundaries are constructed from the cumulative target interval masses as
$
\widetilde e_{d,0}=-1,
\widetilde e_{d,b}
=
-1
+
2\sum_{j=1}^{b}\bar r_{d,j}.
$

For $u_{i,d}\in[e_{b-1},e_b]$, the warped coordinate is obtained by linear interpolation between the corresponding output boundaries:
\begin{equation}
\widetilde u_{i,d}
=
\widetilde e_{d,b-1}
+
\frac{u_{i,d}-e_{b-1}}
{e_b-e_{b-1}}
\left(
\widetilde e_{d,b}
-
\widetilde e_{d,b-1}
\right).
\end{equation}

This design allocates more coordinate capacity to densely sampled intervals while keeping sparse regions closer to uniform allocation. Applying the three axis-wise mappings yields the final spatiotemporal query
$\widetilde{\mathbf q}_i(t)
=(\widetilde u_{i,x},\widetilde u_{i,y},\widetilde u_{i,z},t)$, from which the plane feature is queried as
$\mathbf{g}_{i,p}^{(l)}(t)
=\mathbf{g}_{p}^{(l)}(\widetilde{\mathbf q}_i(t))$.
To prevent the feature query coordinates from changing continuously during optimization, we construct the axis-wise histograms and coordinate mappings once at the end of the coarse stage using the current anchor distribution, and keep them fixed throughout the subsequent fine-stage optimization and anchor densification.

\section{Experiments}

\subsection{Datasets and Implementation Details}

\begin{figure*}[t]
    \centering
    \includegraphics[width=1.0\textwidth]{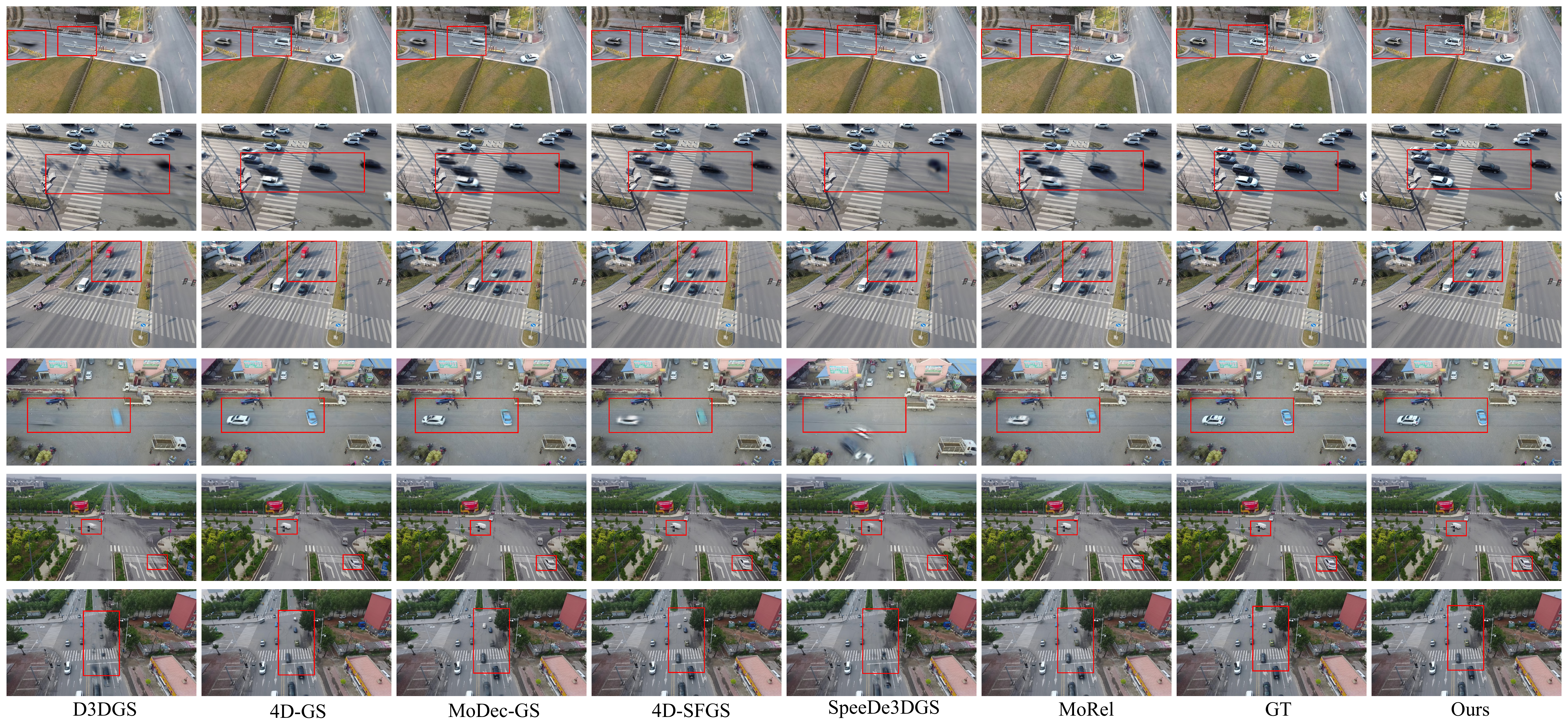}
    \caption{Qualitative comparisons on UAV-Arc4D and VisDrone.
The top three rows correspond to UAV-Arc4D, while the bottom three rows correspond to VisDrone. The red boxes highlight moving objects and challenging motion boundaries.}
    \label{fig:3}
\end{figure*}

\paragraph{Datasets and Evaluation.}
We evaluate our method on three monocular UAV dynamic scene datasets: our self-collected UAV-Arc4D, which contains 10 urban road and traffic scenes, as well as 6 sequences from VisDrone~\cite{zhu2021detection} and 5 sequences from UAVDT~\cite{du2018unmanned}. See the supplementary material for details. For all scenes, we run COLMAP~\cite{schoenberger2016sfm,schoenberger2016mvs} on the complete video sequence to estimate camera poses and generate the initial sparse point cloud.
We evaluate rendering quality using PSNR, SSIM, and LPIPS. We additionally report the speed of the complete rendering pipeline in frames per second (FPS), training time in minutes, and neural representation size (Neural Size, MB). All compared methods use the same camera poses, sparse initialization, training/test split, input images, and main-stage optimization iterations.
All efficiency-related metrics are measured using the same hardware and rendering resolution.

\paragraph{Implementation Details.}
AdaAnchor4D is implemented in PyTorch and optimized on a single NVIDIA RTX 3090 GPU using the Adam optimizer, with 3,000 iterations for the coarse stage and 30,000 iterations for the fine stage. For ACFA, the aggregation embedding dimension and the hidden dimension of the
weight prediction network are set to 32 and 256, respectively. For DLGD, the offset and scale branches employ independent MLPs, with a
geometry condition embedding dimension of 32 and a hidden dimension of 128. For DACW, we construct a density histogram with 64 bins along each spatial axis. Additional configurations and hyperparameters are provided in the supplementary material.

\begin{table*}[t]
    \centering
    \caption{
        Quantitative comparison on UAV-Arc4D and VisDrone.
        Higher PSNR, SSIM, and FPS are better, while lower LPIPS,
        training time, and Neural Size are better.
        Best and second-best results are highlighted in
        bold and underlined, respectively.
    }
    \label{tab:comparison_arc4d_visdrone}

    \setlength{\tabcolsep}{3.5pt}
    \renewcommand{\arraystretch}{1.1}

    \resizebox{\textwidth}{!}{
    \begin{tabular}{llcccccccccccc}
        \toprule
        \multirow{2}{*}{Method}
        & \multirow{2}{*}{Venue}
        & \multicolumn{6}{c}{UAV-Arc4D}
        & \multicolumn{6}{c}{VisDrone} \\
        \cmidrule(lr){3-8}
        \cmidrule(lr){9-14}

        &
        & PSNR$\uparrow$
        & SSIM$\uparrow$
        & LPIPS$\downarrow$
        & FPS$\uparrow$
        & Time$\downarrow$
        & Size$\downarrow$
        & PSNR$\uparrow$
        & SSIM$\uparrow$
        & LPIPS$\downarrow$
        & FPS$\uparrow$
        & Time$\downarrow$
        & Size$\downarrow$ \\
        \midrule

        D3DGS
        & CVPR 2024
        & 29.21
        & 0.929
        & 0.120
        & 18.42
        & 75.70
        & \underline{2.00}
        & 28.95
        & 0.900
        & 0.149
        & 14.32
        & 126.17
        & \underline{2.00} \\

        4D-GS
        & CVPR 2024
        & 31.90
        & 0.922
        & 0.149
        & \underline{52.45}
        & 45.60
        & 27.30
        & 30.00
        & 0.870
        & 0.210
        & \underline{42.12}
        & \textbf{56.83}
        & 27.50 \\

        MoDec-GS
        & CVPR 2025
        & \underline{32.05}
        & \underline{0.937}
        & 0.114
        & 33.85
        & 73.80
        & 30.00
        & \underline{30.71}
        & \underline{0.913}
        & 0.133
        & 24.33
        & 105.00
        & 30.50 \\

        4D-SFGS
        & AAAI 2026
        & 31.67
        & \underline{0.937}
        & \underline{0.109}
        & 2.30
        & \textbf{39.70}
        & \textbf{0.12}
        & 30.43
        & 0.910
        & \underline{0.127}
        & 1.79
        & 87.67
        & \textbf{0.11} \\

        SpeeDe3DGS
        & CVPR 2026
        & 28.49
        & 0.911
        & 0.165
        & \textbf{133.42}
        & 47.00
        & \underline{2.00}
        & 28.16
        & 0.869
        & 0.216
        & \textbf{98.86}
        & 70.83
        & \underline{2.00} \\

        MoRel
        & CVPR 2026
        & 29.68
        & 0.930
        & 0.115
        & 24.51
        & 72.50
        & 17.00
        & 29.25
        & 0.907
        & 0.132
        & 10.71
        & 86.00
        & 19.50 \\

        \midrule

        \textbf{Ours}
        & --
        & \textbf{33.74}
        & \textbf{0.943}
        & \textbf{0.102}
        & 47.04
        & \underline{43.29}
        & 9.00
        & \textbf{31.91}
        & \textbf{0.921}
        & \textbf{0.118}
        & 32.41
        & \underline{57.60}
        & 7.60 \\

        \bottomrule
    \end{tabular}
    }
\end{table*}

\subsection{Comparisons}

We compare AdaAnchor4D with several representative dynamic Gaussian reconstruction methods, including D3DGS~\cite{yang2024deformable}, 4D-GS~\cite{wu20244d}, MoDec-GS~\cite{kwak2025modec}, 4D-SFGS~\cite{cho20264d}, SpeeDe3DGS~\cite{TuYing2025SpeeDe3DGS}, and MoRel~\cite{kwak2026morel}.

\paragraph{Reconstruction Quality.}
As shown in Tables~\ref{tab:comparison_arc4d_visdrone}
and~\ref{tab:comparison_uavdt}, AdaAnchor4D achieves the best rendering quality on UAV-Arc4D, VisDrone, and UAVDT. On UAV-Arc4D, our method achieves a PSNR of 33.74 dB, an SSIM of 0.943, and an LPIPS of 0.102, outperforming the second-best results by 1.69 dB in PSNR and 0.006 in SSIM, while reducing LPIPS by 0.007. On VisDrone, AdaAnchor4D achieves a PSNR of 31.91 dB, an SSIM of 0.921, and an LPIPS of 0.118, improving PSNR and SSIM by 1.20 dB and 0.008, respectively, while reducing LPIPS by 0.009. On UAVDT, our method achieves a PSNR of 29.51 dB, an SSIM of 0.909, and an LPIPS of 0.101, surpassing the second-best results by 0.59 dB in PSNR and 0.006 in SSIM, with a 0.004 reduction in LPIPS. These results demonstrate that AdaAnchor4D consistently improves rendering quality across all three dynamic UAV-scene datasets.

\paragraph{Qualitative Comparison.}
As shown in Fig.~\ref{fig:3}, we compare AdaAnchor4D with representative methods on dynamic UAV scenes from UAV-Arc4D and VisDrone. The red boxes highlight moving vehicles and challenging motion boundaries. Existing methods often produce ghosting artifacts, blurred contours, and local structural distortions in regions with complex motion. In contrast, AdaAnchor4D recovers sharper object boundaries and more coherent local geometry and appearance, producing results that are closer to the ground truth. Additional quantitative results and qualitative visualizations are provided in the supplementary material.

\begin{table}[t]
    \centering
    \caption{
        Quantitative comparison on UAVDT.
        Best and second-best results are highlighted in
        bold and underlined, respectively.
    }
    \label{tab:comparison_uavdt}

    \setlength{\tabcolsep}{3.5pt}
    \renewcommand{\arraystretch}{1.1}

    \resizebox{\columnwidth}{!}{
    \begin{tabular}{lcccccc}
        \toprule
        Method
        & PSNR$\uparrow$
        & SSIM$\uparrow$
        & LPIPS$\downarrow$
        & FPS$\uparrow$
        & Time$\downarrow$
        & Size$\downarrow$ \\
        \midrule

        D3DGS
        & 26.83
        & 0.896
        & 0.126
        & 8.25
        & 150.23
        & \underline{2.00} \\

        4D-GS
        & 27.52
        & 0.836
        & 0.218
        & 31.65
        & \textbf{63.18}
        & 28.60 \\

        MoDec-GS
        & 28.22
        & 0.898
        & 0.123
        & 23.45
        & 141.60
        & 30.20 \\

        4D-SFGS
        & \underline{28.92}
        & \underline{0.903}
        & \underline{0.105}
        & 2.14
        & 93.00
        & \textbf{0.11} \\

        SpeeDe3DGS
        & 26.12
        & 0.849
        & 0.207
        & \textbf{104.44}
        & 68.80
        & \underline{2.00} \\

        MoRel
        & 26.21
        & 0.871
        & 0.145
        & 11.52
        & 115.40
        & 26.60 \\

        \midrule

        \textbf{Ours}
        & \textbf{29.51}
        & \textbf{0.909}
        & \textbf{0.101}
        & \underline{32.94}
        & \underline{67.21}
        & 9.40 \\

        \bottomrule
    \end{tabular}
    }
\end{table}

\paragraph{Efficiency Analysis.}
AdaAnchor4D maintains real-time rendering performance, achieving 47.04, 32.41, and 32.94 FPS on UAV-Arc4D, VisDrone, and UAVDT, respectively, with corresponding training times of 43.29, 57.60, and 67.21 minutes. Although SpeeDe3DGS achieves higher rendering speeds and some competing methods require less neural representation storage, their rendering quality remains lower than that of AdaAnchor4D. The Neural Size of our method is 9.0, 7.6, and 9.4 MB on the three datasets, respectively, indicating modest neural representation storage. Overall, AdaAnchor4D achieves a favorable quality--efficiency trade-off by substantially improving rendering quality while retaining real-time rendering capability and a compact neural representation.

\begin{figure}[t]
    \centering
    \includegraphics[width=0.95\linewidth]{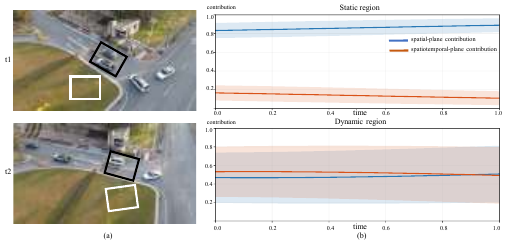}
    \caption{Visualization of the learned ACFA aggregation weights.
    (a) Ground-truth images at two different time steps, where the white and black boxes indicate representative static and dynamic regions, respectively.
    (b) Temporal variation of the normalized spatial-plane and spatiotemporal-plane contributions of anchors within the corresponding regions. Solid lines denote the mean across anchors, and shaded regions indicate one standard deviation.}
    \label{fig:4}
\end{figure}

\subsection{Ablation Studies}

\paragraph{Module Ablation.}
Table~\ref{tab:ablation_arc4d_visdrone} presents the ablation studies of ACFA, DLGD, and DACW on UAV-Arc4D and VisDrone. The baseline adopts the same anchor-based dynamic framework, aggregates shared spatiotemporal features using Hadamard products, models only the temporal variations of anchor positions and features, and employs linear coordinate normalization.

Introducing ACFA improves PSNR from 32.18/30.68 dB to 33.16/31.29 dB on UAV-Arc4D/VisDrone, while reducing LPIPS from 0.111/0.130 to 0.106/0.126. Building upon ACFA, incorporating DLGD further increases PSNR to 33.52/31.51 dB, whereas incorporating DACW achieves 33.54/31.66 dB, with an SSIM of 0.943/0.919 and an LPIPS of 0.104/0.124. Combining all three modules yields the best overall performance, achieving 33.74/31.91 dB PSNR, 0.943/0.921 SSIM, and 0.102/0.118 LPIPS. Compared with the baseline, the full model improves PSNR by 1.56/1.23 dB on the two datasets. These results demonstrate the effectiveness of ACFA, DLGD, and DACW for
adaptive feature aggregation, decoupled local geometry deformation, and density-adaptive coordinate warping, respectively, as well as their complementary benefits.

\paragraph{Feature Aggregation.}
As shown in Table~\ref{tab:ablation_acfa_fusion}, Product and Mean achieve similar performance on both datasets, indicating that simply replacing the fixed aggregation operator provides little benefit. Anchor-only improves the results on UAV-Arc4D but shows no gain on VisDrone, whereas Time-only mainly benefits VisDrone, suggesting that either conditioning source alone is insufficient for consistent improvement. In contrast, ACFA jointly incorporates anchor-specific aggregation embeddings and temporal information and achieves the best performance across all metrics. Compared with Product, ACFA improves PSNR by 0.98/0.61 dB and reduces LPIPS by 0.005/0.004 on UAV-Arc4D/VisDrone. These results show that ACFA benefits from jointly conditioning plane-feature aggregation on anchor-specific aggregation embeddings and temporal information.
To further validate the adaptivity of ACFA, Fig.~\ref{fig:4} visualizes the learned aggregation weights in representative static and dynamic regions. Static regions rely more heavily on spatial planes, whereas dynamic regions assign higher average weights to spatiotemporal planes, indicating that ACFA adaptively adjusts shared feature aggregation according to local dynamic states.

\begin{table}[t]
\centering
\caption{Ablation results on UAV-Arc4D and VisDrone.
The best results are highlighted in bold.}
\label{tab:ablation_arc4d_visdrone}
\setlength{\tabcolsep}{2.5pt}
\renewcommand{\arraystretch}{1.08}

\resizebox{\columnwidth}{!}{
\begin{tabular}{ccccccccc}
    \toprule
    ACFA
    & DLGD
    & DACW
    & \multicolumn{3}{c}{UAV-Arc4D}
    & \multicolumn{3}{c}{VisDrone} \\
    \cmidrule(lr){4-6}
    \cmidrule(lr){7-9}

    &
    &
    &
    PSNR$\uparrow$
    & SSIM$\uparrow$
    & LPIPS$\downarrow$
    & PSNR$\uparrow$
    & SSIM$\uparrow$
    & LPIPS$\downarrow$ \\
    \midrule

    &
    &
    &
    32.18
    & 0.937
    & 0.111
    & 30.68
    & 0.913
    & 0.130 \\

    \cmark
    &
    &
    &
    33.16
    & 0.941
    & 0.106
    & 31.29
    & 0.916
    & 0.126 \\

    \cmark
    & \cmark
    &
    &
    33.52
    & \textbf{0.943}
    & 0.104
    & 31.51
    & 0.917
    & 0.120 \\

    \cmark
    &
    & \cmark
    &
    33.54
    & \textbf{0.943}
    & 0.104
    & 31.66
    & 0.919
    & 0.124 \\

    \cmark
    & \cmark
    & \cmark
    &
    \textbf{33.74}
    & \textbf{0.943}
    & \textbf{0.102}
    & \textbf{31.91}
    & \textbf{0.921}
    & \textbf{0.118} \\

    \bottomrule
\end{tabular}
}
\end{table}
\begin{table}[t]
\centering
\caption{
Ablation study of feature aggregation and conditioning strategies.
All variants deform only the anchor position and feature while using
the same shared spatiotemporal feature field and training configuration.
}
\label{tab:ablation_acfa_fusion}
\setlength{\tabcolsep}{2.5pt}
\renewcommand{\arraystretch}{1.05}

\resizebox{\columnwidth}{!}{
\begin{tabular}{lcccccc}
    \toprule
    Strategy
    & \multicolumn{3}{c}{UAV-Arc4D}
    & \multicolumn{3}{c}{VisDrone} \\
    \cmidrule(lr){2-4}
    \cmidrule(lr){5-7}

    &
    PSNR$\uparrow$
    & SSIM$\uparrow$
    & LPIPS$\downarrow$
    & PSNR$\uparrow$
    & SSIM$\uparrow$
    & LPIPS$\downarrow$ \\
    \midrule

    Product
    & 32.18
    & 0.937
    & 0.111
    & 30.68
    & 0.913
    & 0.130 \\

    Mean
    & 32.14
    & 0.937
    & 0.111
    & 30.69
    & 0.912
    & 0.131 \\

    Anchor-only
    & 32.53
    & 0.938
    & 0.110
    & 30.67
    & 0.913
    & 0.131 \\

    Time-only
    & 32.19
    & 0.938
    & 0.112
    & 30.91
    & 0.914
    & 0.130 \\

    ACFA
    & \textbf{33.16}
    & \textbf{0.941}
    & \textbf{0.106}
    & \textbf{31.29}
    & \textbf{0.916}
    & \textbf{0.126} \\

    \bottomrule
\end{tabular}
}
\end{table}

\section{Conclusion}
We presented AdaAnchor4D, an adaptive anchor deformation
framework for monocular UAV 4D reconstruction. ACFA
conditions the aggregation of a decomposed shared
spatiotemporal field on anchor-specific aggregation
embeddings and temporal information, enabling local units
to obtain adaptive dynamic representations from the same
compact shared field. DLGD further separates anchor-state
deformation from local Gaussian geometry deformation,
while DACW reparameterizes feature-query coordinates
according to the axis-wise anchor distributions. Experiments
on three UAV dynamic-scene datasets demonstrate that
AdaAnchor4D improves rendering quality while maintaining real-time rendering performance and a
compact neural representation. Our framework relies on
pre-estimated camera poses and may be less robust to severe
pose errors, occlusions, and sparse observations. Future work
will jointly refine camera poses and dynamic scene
representations.

\bibliography{aaai2027}
\end{document}